\documentclass[letterpaper]{article} 
\usepackage{aaai24}  
\usepackage{times}  
\usepackage{helvet}  
\usepackage{courier}  
\usepackage[hyphens]{url}  
\usepackage{multirow}
\usepackage{graphicx} 
\usepackage{natbib}  
\usepackage{caption} 
\usepackage{colortbl}
\usepackage{algorithm}
\usepackage{algorithmic}
\usepackage{amsmath}
\usepackage{amssymb}
\usepackage{newfloat}
\usepackage{listings}

\usepackage{xcolor} 
\usepackage{subcaption}
\usepackage{array}
\newcommand{\jy}[1]{{\color{black}#1}}

\definecolor{c1}{HTML}{ff0097}

\DeclareCaptionStyle{ruled}{labelfont=normalfont,labelsep=colon,strut=off} 
\floatstyle{ruled}
\newfloat{listing}{tb}{lst}{}
\floatname{listing}{Listing}
\title{

Relax Image-Specific Prompt Requirement in SAM: A Single Generic Prompt for Segmenting Camouflaged Objects
}

\author{
    Jian Hu\textsuperscript{\rm 1}\equalcontrib,
    Jiayi Lin\textsuperscript{\rm 1}\equalcontrib,
    Weitong Cai\textsuperscript{\rm 1},
    Shaogang Gong\textsuperscript{\rm 1}
}
\affiliations{
    \textsuperscript{\rm 1}Queen Mary University of London\\
    \{jian.hu, 
    jiayi.lin,
    weitong.cai,
    s.gong\}@qmul.ac.uk\\
    \large{\textcolor{c1}{https://lwpyh.github.io/GenSAM/}}
}

\usepackage{bibentry}
\begin{document}

\maketitle
\begin{figure*}[ht]
  \centering
  \begin{tabular}{@{\hspace{0pt}} c @{\hspace{5pt}}  c }
    {\includegraphics[width=1.4\columnwidth]{
    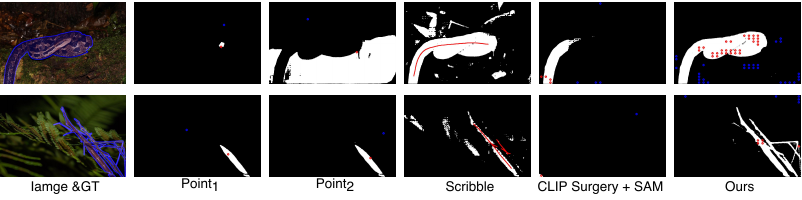}}&
      {\includegraphics[width=0.55\columnwidth]{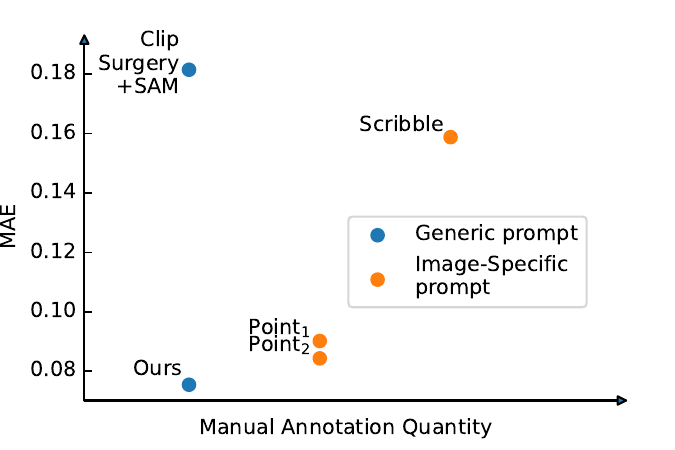}} \\[-0.8ex]
    {\small (a) Segmentation results using different prompts in SAM with various approaches.} &
     { \small (b) Mean absolute error on S-COD dataset.}
  \end{tabular}
  \caption{In SAM, manual point and scribble prompts suffer from  ambiguity in interpreting targets and is sensitive to minor spatial variations. 
  Using a generic task description as a generic prompt with CLIP Surgery+SAM enables the model to achieve some segmentation capability on obvious objects. However, it struggles to perform well in complex environments with camouflage-like patterns. In contrast, our proposed GenSAM can adaptively convert a generic task description into image-specific visual prompts, effectively enhancing the segmentation process by leveraging the unique characteristics of each image. 
  \label{fig:moti}}
\end{figure*}
\begin{abstract}
Camouflaged object detection (COD) approaches heavily rely on pixel-level annotated datasets. 
Weakly-supervised COD (WSCOD) approaches use sparse annotations like scribbles or points to reduce annotation efforts, but this can lead to decreased accuracy. 
The Segment Anything Model (SAM) shows remarkable segmentation ability with sparse prompts like points. 
However, manual prompt is not always feasible, as it may not be accessible in real-world application. 
Additionally, it only provides localization information instead of semantic one, which
can intrinsically cause ambiguity in interpreting targets.
In this work, we aim to eliminate the need for manual prompt.
The key idea is 
to employ Cross-modal Chains of Thought Prompting (CCTP) to reason visual prompts using the semantic information given by a generic text prompt.
To that end, we introduce a test-time instance-wise adaptation mechanism 
called Generalizable SAM (GenSAM) to automatically generate and optimize visual prompts 
from the generic task prompt for WSCOD.
In particular, CCTP maps a single generic text prompt 
onto image-specific consensus foreground and background heatmaps
using vision-language models, 
acquiring reliable visual prompts.
Moreover, to test-time adapt the visual prompts, 
we further propose Progressive Mask Generation (PMG) to iteratively reweight the input image, 
guiding the model to focus on the targeted region in a coarse-to-fine manner.
Crucially, all network parameters are fixed, avoiding the need for additional training.
%
Experiments on three benchmarks demonstrate that GenSAM outperforms point supervision approaches and achieves comparable results to scribble supervision ones, solely relying on general task descriptions.
\end{abstract}

\section{Introduction}
\label{Intro}
Camouflaged Object Detection (COD) aims to accurately identify inconspicuous objects that have been carefully disguised, including those found in natural and artificial environments \cite{fan2017structure}.
The task's complexity is amplified by the indistinct boundaries between objects and backgrounds, necessitating a significant number of precisely annotated image-mask pairs. This places a rigorous demand on the annotation process \cite{hubel1962receptive, perez2012early, pang2022zoom}.
To alleviate this burden, weakly-supervised COD (WSCOD) is introduced to relax the annotation requirements.
That only requires a sparse annotation in either the foreground or background.
However, as annotations become sparser, they suffer from reduced accuracy.
SAM introduces instance-level prompts to optimize segmentation, promising performance can be achieved with only a few points as prompts for each instance.

However, SAM has limited comprehension on its segmented object.
Manual prompts can only provide location information of desired segmentation objects, but lack in semantic information, leading to potential ambiguity. 
In Fig.~\ref{fig:moti}(a), despite both prompts targeting the same object, minor changes in the point prompt's position can make SAM misinterpret the desired object, greatly changing the results.
Moreover, compared to human perception, it also exhibits a strong bias in interpreting prompts~\cite{chen2023ability}. 
In Fig.~\ref{fig:moti}(b), even with more information from scribble prompts than point prompts, SAM still misunderstands the target, leading to limited performance.
To eliminate ambiguity and bias, recent work expands manual prompt input options to include reference regions, videos and even audios~\cite{zou2023segment}. 
However, regardless of the prompting method, 
SAM still requires instance-specific manual prompts, 
which may not always be practical in real-world scenarios, 
and the question of eliminating this need remains unexplored.

In this work, we introduce a test-time adaptation mechanism called Generalizable SAM (GenSAM), a novel approach to alleviating the demand for accurate, instance-specific manual prompts in the SAM framework for the WSCOD task.
Given a simple text description, detailed semantic information about the desired object is reasoned based on both text and image information. 
Subsequently, this generates unambiguous visual prompt to guide the segmentation without human intervention.
In order to provide semantic information of the target objects for SAM, we introduce Cross-modal Chains of Thought Prompting (CCTP) which automatically reason pixel-level visual prompts from various chains. 
BLIP2~\cite{zhu2023chatgpt} and our devised CLIP~\cite{radford2021learning} are employed for this propose.
%
BLIP2 generates various keywords related to the target and their background using multiple chains of prompting~\cite{wei2022chain},
ambiguously in the input text prompts is eliminated. 
The spatial CLIP component introduces a novel self-attention mechanism, mapping keywords from different chains onto a consensus heatmap for generating consensus visual prompts, thus resolving the visual prompt ambiguity induced by a particular chain.
To progressively adapt the visual prompts, we further propose Progressive Mask Generation (PMG), which uses an test-time prompt tuning approach to iteratively reweight the input image with the consensus heatmaps. This guides the model to focus on the targets in a  coarse-to-fine manner.
It encourages the model to concentrate on task-relevant regions, enhancing the performance.

\textbf{Our contribution can be summarized as follows:}

\noindent(1) To eliminate the need for specific annotations tailored to each image in WSCOD, our GenSAM approach automatically generates personalized prompts for multiple unlabeled images using only a general task description.
%

\noindent(2) To convert task descriptions into precise visual prompts, we introduce an  Cross-modal Chains of Thought Prompting module. It uses a consensus mechanism and a novel self-attention to derive image-specific prompts for SAM.
Additionally, Progressive Mask Generation module utilizes the consensus heatmap as a visual prompt, progressively enhancing the segmentation performance.

\noindent(3) Extensive experiments on three benchmarks have demonstrated the effectiveness of our proposed GenSAM.
 \begin{figure*}[tb!]
   \centering   \includegraphics[width=17.8cm]{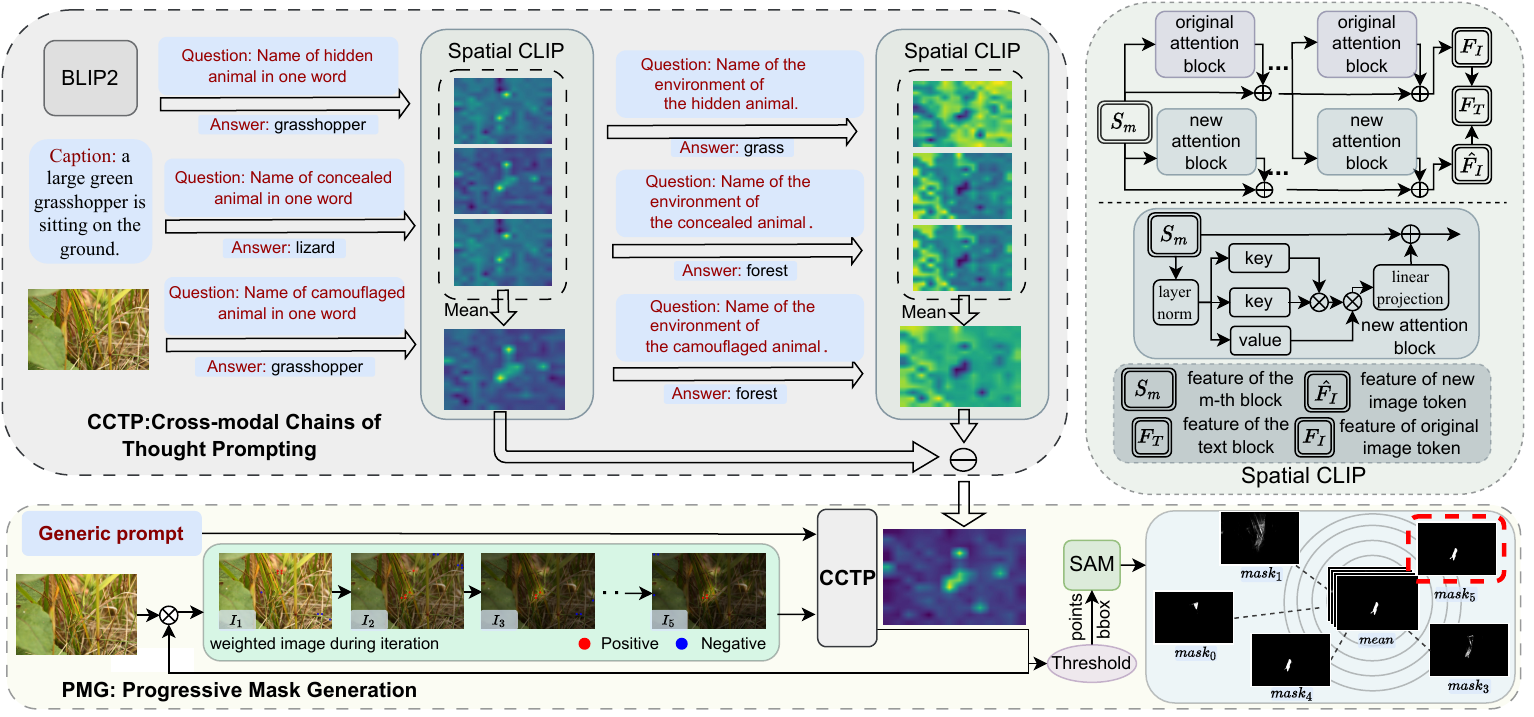}
   \caption{The framework of our proposed GenSAM.   GenSAM consists of two components: Cross-modal Chains of Thought Prompting (CCTP) and Progressive Mask Generation (PMG). 
   CCTP begins by taking a generic task prompt as input. BLIP2 generates an image caption for each image, using the input generic prompt as a foundation. 
   Based on this prompt and generated caption, three parallel chains of thought are constructed to extract keywords about concealed animals and their corresponding background from unlabeled images. 
   These keywords are then fed into our designed spatial CLIP module, which generates heatmaps for locating the camouflaged objects. High-confidence regions selected from these heatmaps serve as prompts to guide the segmentation process. 
   The heatmaps generated by CCTP are weighted and utilized as visual prompts in PMG, gradually directing the model's attention towards task-relevant regions. 
   In addition, during the adaptation process, the mask generated by a single iteration that is closest to the average mask obtained from multiple iterations is selected as the final output.
   \label{fig:framework}}
\end{figure*}
\section{Related Work}
 \subsection{Concealed Object Segmentation}
 The goal of camouflage detection is to identify objects that blend into complex backgrounds. 
 Initially, some studies utilize low-level features such as texture, brightness, and color to distinguish the foreground from the background \cite{pike2018quantifying, hou2011detection, sengottuvelan2008performance}. 
 Recently, several end-to-end approaches \cite{fan2020pranet, fan2020camouflaged} are proposed that achieve superior performance. However, most of these methods require fully annotated samples for training, which imposes a significant annotation burden.
Weakly-Supervised Concealed Object Detection (WSCOD) aims to train a segmentation model using sparsely annotation like points and scribbles.
Although WSCOD alleviates the reliance on pixel-level annotations, its performance is still limited by the quality and diversity of the training data. 
The lack of representative samples in a single dataset, as well as the restricted coverage of various scenes and objects, hinder the generalization ability of the model.
For example, \cite{he2023weakly} only requires scribble supervision, which achieves decent segmentation performance with lower annotation requirements per image. 
But its performance is limited by the quality of the annotations and lacks in generalization ability. 
%
%
Furthermore, to achieve satisfied performance across different datasets, current WSCOD approaches still requires separate training on different datasets, which limits their generalization ability.
 \subsection{Segment Anything model}
Segment Anything Model (SAM) \cite{kirillov2023segment} is trained on the extensive SA-1B dataset, which comprises a vast collection of 11 million images and over 1 billion masks. This extensive dataset enables SAM to establish a robust foundation model for image segmentation with strong zero-shot generalization ability.
While SAM is good at segmenting images, it struggles with segmenting camouflaged objects \cite{tang2023can, ji2023sam, ji2023segment}. 
Moreover, its impressive ability requires the use of carefully crafted prompts to guide segmentation, which can be subjective and unclear.
To address the challenges SAM encounters in camouflaged object detection, SAM-adaptor \cite{chen2023sam} leverages a fully supervised dataset of camouflaged objects to train the encoder, yielding favorable results. However, this approach is hampered by its substantial demand for pixel-level annotated data. 
On the other hand, PLFMG \cite{he2023weakly} enhances SAM's performance in WSCOD task through the application of pseudo labeling and multi-scale feature grouping. Regrettably, this method remains contingent upon separate training for different datasets within the WSCOD task, indicative of a deficiency in robust generalization capabilities.
In contrast, our proposed approach only mandates a generic task description, enabling us to perform effective segmentation of concealed objects in unsupervised images across diverse datasets within the WSCOD task through instance-level test-time adaptation.

\subsection{Test-time Adaptation}
Test-time domain adaptation aims to adapt the model to a test domain that exhibits a domain gap with the training data~\cite{wang2020tent, hu2020discriminative}, in order to improve the performance on the test data~\cite{niu2022efficient}. 
Currently, there are two main categories of test-time domain adaptation: backward-based adaptation and backward-free adaptation. 
The former often utilizes self-supervised learning methods to learn the data characteristics of the target domain through entropy minimization \cite{wang2020tent,hu2019multi,hu2022learning}. 
The latter mostly achieves backward-free adaptation through batch normalization statistic adaptation.
DUA \cite{mirza2022norm} employs a running average technique to update the statistics and achieve adaptation, while DIGA \cite{wang2023dynamically} utilizes distribution adaptation via batch normalization to effectively perform semantic segmentation under domain gaps. 
In our work, we employ instance-level test-time domain adaptation, which simply relies on a general task description for camouflage object segmentation. It allows accurate camouflage object segmentation across diverse datasets without sample-level supervision.
\section{Methodology}
We present GenSAM for segmenting camouflaged objects among different domains, based on a general task description.
%
In specific, we (1) propose \emph{Cross-modal Chains of Thought Prompting}, which reasons the description of targeted objects in each image and further derives a consensus attention heatmap to generate visual prompts for the SAM model, and (2)
 employ an iterative process \emph{Progressive Mask Generation} to apply the consensus heatmap onto the original image as a visual prompt to further improve  segmentation results.Note that GenSAM is entirely training-free, only relying on pretrained components without additional training data or extra parameters during test-time  adaptation.
 
Given an image $X \in \mathbb{R}^{H \times W \times 3}$ from a test set and a task-generic prompt $P_g$ (``the camouflaged animal''), GenSAM aim at inferring the visual prompts for SAM to get the final segmentation mask $M \in \mathbb{R}^{H \times W}$.
We relax the requirement for each unlabeled image under the same task to have a unique supervision and instead, adopt a common task-generic prompt shared by different unlabeled images across datasets within the same task.
%
%
%

%
\subsection{Cross-modal Chains of Thought Prompting}
The task of converting task-generic text prompts into image-specific visual prompts poses two main challenges: generating robust and objective image-specific prompts, and accurately localizing camouflaged objects in the images for effective segmentation.
%
Therefore, we use multiple cross-modal chains of thought to evaluate unlabeled images from different perspectives, generating potential keywords for both the camouflaged objects and their backgrounds.
These keywords are then fed into our spatial CLIP module, which generates specific heatmaps for each foreground and background keyword.
The foreground and background heatmaps undergo individual consensus calculation, and the background consensus heatmap is subtracted from the foreground consensus heatmap. The resulting heatmap highlights regions of high confidence, which are selected as prompts and used for segmentation with the SAM model.

\noindent \textbf{Keyword Generation with various chains of thought.}
We utilize multiple chains of thought to produce a variety of keywords for both interested objects and their background, and generate heatmaps for both to remove irrelevant highlights of the background in the heatmaps.
Specifically, we utilize BLIP2, an image-to-caption model that generates task-relevant object keywords based on generic prompts. These keywords are used to aid in localizing the objects of interest.
 However, directly using the task-generic prompt $P_g$ to query BLIP2 for camouflaged objects in the image leads to inaccurate answers (Tab.~\ref{table:component}). 
 Inspired by generated knowledge prompting \cite{liu2021generated}, we propose a method that involves having BLIP2 initially generate a caption $C$ for the image $X$. 
 This generated caption is then incorporated to enhance the model's ability to make more precise predictions when querying about the image-specific targets.
\begin{equation}
    C=BLIP2(X),
\end{equation}

As in large-scale generative models (LLM) \cite{bai2021training, zhu2023chatgpt} like BLIP2, prompts require careful design and even slight variations in the prompts can lead to significant differences in the generated keywords.
\cite{wang2022self} proposes to design various chains of prompts for generative language models, and then derive a consensus from the output results to serve as the final output.
It assumes that the output results of BLIP2 can be determined through majority voting among a limited number of possible outcomes.
Therefore, a single prompt often provides only partial and biased descriptions of objects. 
Therefore, for the same image $X$, we inquire from different perspectives using various prompts to obtain different descriptive keywords ${A}^{fore}_j$ for interested foreground objects (camouflaged objects), where ${fore}$ represents the foreword keyword and $j$ denotes the $j-$th chain of thought. 
As shown in Fig. \ref{fig:framework}, with the corresponding task description $P_g$ as ``camouflaged animal'' we first have BLIP2 generate a caption $C$ for the image $X$. 
Then, using $C$ as a basis, we replace $P_g$ ``camouflaged animal'' with two synonymous phrases, ``hidden animal'' and ``concealed animal'' creating  $J$ different chains of thought $\{(Q_j^{fore},Q_j^{back})\}_{j=1}^{J}$ with similar meanings from different perspective, simultaneously propose different questions in parallel to inquire about $X$.
For example, $Q_1^{fore}$:``Name of the hidden animal in one word'', $Q_2^{fore}$:``Name of the concealed animal in one word'' and $Q_3^{fore}$:``Name of the camouflaged animal in one word.''.
Then, the obtained foreground keyword $A_j^{fore}$ can be denoted as follows,
\begin{equation}
    A_j^{fore}=BLIP2(X, C, Q_j^{fore}),
\end{equation}
%
Moreover, camouflaged objects often hide themselves using textures or backgrounds. Therefore, identifying the background of camouflaged objects can significantly mitigate interference from unrelated objects.
%
%
To achieve this, we further use the background query $Q_j^{back}$ by including inquiries about the background of the $A_j^{fore}$ (e.g, ``grasshopper'' in Fig. \ref{fig:framework}).
This enables us to obtain the background keywords $A_j^{back}$ as follows,
\begin{equation}
    A_j^{back}=BLIP2(X, C, Q_j^{fore}, A_j^{fore}, Q_j^{back}).
\end{equation}

\subsubsection{Spatial CLIP.}
Due to CLIP's powerful open-vocabulary capability, it can handle various text descriptions. 
Hence, we input the generated keywords ${A}_j^{fore}$ and $A_j^{back}$ into CLIP, aiming to leverage its cross-modal alignment capability to highlight the corresponding regions in the image related to the interested object. 
However, CLIP's open-vocabulary capability also results in the generated heatmap containing numerous irrelevant information unrelated to the task.

Clip Surgery \cite{li2023clip} employs v-v-v self-attention to address it, where the queries (Q), keys (K), and values (V) in self-attention mechanism are all replaced by the values (V). 
The similarity between queries and keys is computed using the same vectors. 
This design enhances computational efficiency as there is no need to differentiate between queries and keys. 
But using the same representation for queries, keys, and values might limit the model's ability to capture internal correlations and features within the input image token, as they are mixed in the representation space.

To further enhance the location accuracy of the heatmap and effectively explore the internal structures and semantic correlations within the image, we propose the k-k-v self-attention paralleled to the original k-q-v path.
The element-wise multiplication of the keys vectors reduces interference from redundant features, enabling the self-attention mechanism to focus more on internal correlations within the input image. 
This results in a better representation of the image's internal structure and patterns. Additionally, the ``kkv'' approach, using different vectors for values, preserves more information from the original inputs, enriching the context and enhancing the model's expressive capabilities.
For the $m$-th block of the visual encoder in CLIP, its input features are denoted as $S_m$, where $S_m \in \mathbb{R}^{L \times d}$, with $K=S_m \cdot W_k$, $V=S_m \cdot W_v$, and $Q=S_m \cdot W_q$.
Here, $W_k \in \mathbb{R}^{d \times d_k}$, $W_v \in \mathbb{R}^{d \times d_v}$, and $W_q \in \mathbb{R}^{d \times d_q}$ are learnable parameter matrices.
Next, we split $K$, $V$, and $Q$ into $h_0$ individual heads, each with dimensions $d_k$, $d_v$, and $d_q$, respectively. Assuming we have $h_0$ heads, the dimensions of $K$, $V$, and $Q$ can be represented as follows: $K=\{k_1, k_2, ..., k_{h_0}\}$, $Q=\{q_1, q_2, ..., q_{h_0}\}$, and $V=\{v_1, v_2, ..., v_{h_0}\}$. Here, $k_n \in \mathbb{R}^{L \times d_k}$, $v_n \in \mathbb{R}^{L \times d_v}$, and $q_n \in \mathbb{R}^{L \times d_q}$ represent the transformation results of the $n$-th head.
%
In our proposed k-k-v self-attention strategy, Q is replaced by K, and the expression for k-k-v self-attention is as follows:
\begin{equation}
\begin{split}
\text{attention}_{kkv}=\text{softmax}(K*K^T*scale)*V,
\end{split}
\end{equation}
%
where $scale$ is the scaling factor. The output of the K-Q-V block ${S}_{m+1}$ and modified K-K-V self attention block $\hat{S}_{m+1}$ are defined as follows:
\begin{small}
\begin{equation}
\begin{split}
\label{eq:loss_1}
   &S_{m+1} = f_{\text{FFN}}(\text{attention}_{kqv}(S_{m}) + S_{m}),
    \\&\hat{S}_{m+1} = \begin{cases}
    \text{None}, \text{if } m < \delta \\
    f_{\text{FFN}}({\text{attention}_{kkv}(S_{m})}+ S_{m}) 
    \quad \text{if } m = \delta \\
    f_{\text{FFN}}({\text{attention}_{kkv}({S_{m})}}+ \hat{S}_{m}) 
    \quad \text{if } m > \delta
    \end{cases}, 
 \end{split}
\end{equation}
\end{small}
where $\hat{S}_{m}$ and $S_{m}$ represent the output of k-k-v self-attention and k-q-v self-attention, respectively, for the m-th layer block. 
Shallow layers in the spatial CLIP
focus more on low-level features and less on higher level concepts
(e.g. meanings of input keywords). The original k-q-v attention mechanism is used in these layers. 
Deep layers start to represent higher level semantics like people and animals. 
Thus, for layers at depth $\delta$, $S_{m-1}$ is utilized as the input for the new module. Beyond depth $\delta$, $\hat{S}_{m-1}$ and ${S}_{m-1}$from the previous block are both used as input.
Following Clip Surgery~\cite{li2023clip}, the value of $\delta$ is set to 7.
%
%
These features are then accumulated and mixed using the fully connected layer $f_{\text{FFN}}$.
Image $X$, processed through the original image path and the k-k-v image path, yields image features $F_I$ and $\hat{F}_I$, respectively. The corresponding text feature of the keyword $A^{fore}_j$ and $A^{back}_j$ are $F^{fore}_j$ and $F^{back}_j$.

\subsubsection{Visual prompt with consensus.} 
After obtaining the corresponding features of different keywords,
our objective is to identify a consensus among these features to locate specific regions of interest related to the task. 
Specifically, the consensus derived from foreground keyword generation is utilized to pinpoint the precise location of the camouflaged objects, while the consensus derived from background keyword generation helps eliminate interference from the background in object localization.
For foreground keywords, the feature dimensions of the foreground keyword $F^{fore}_j$ and the image feature $\hat{F_I}$ are 
$N_i \times 1 \times C$, where $N_i$ represents the size of the image tokens, 1 represents the size of the text token, 
and $C$ represents the number of channels. 
Consequently, the foreground similarity vector $SI_{fore}^j$ obtained for the $j$-th keyword, is defined as:

\begin{equation}
SI_{fore}^j = \frac{F^{fore}_j}{||F^{fore}_j||_2}
\odot
\frac{\hat{F_I}}{||\hat{F_I}||_2},
\end{equation}
where $\odot$ is element-wise multiplication.
L2 normalization is applied across the channel dimension.
To derive a consensus among various image features corresponding to different foreground keywords, $SI_{fore}$ can be obtained as follows:
\begin{equation}
    SI_{fore} = \frac{\sum_{j=1}^{J}(SI_{fore}^j)}{J},
\end{equation}
where $j$ is the number of the chains, we set $J$ as 3.
\begin{table*}[htbp]
\setlength{\tabcolsep}{6pt}
\small
\centering
 \caption{Results on COD with point supervision and scribble supervision. Best are in \textbf{bold} 
.}
\label{tab:results}
 \resizebox{1.0\textwidth}{!}{
\begin{tabular}{c|c|cccc|cccc|cccc}
\hline
{\multirow{2}{*}{Methods}} & {\multirow{2}{*}{Venue}} &\multicolumn{4}{c}{
CHAMELEON}&\multicolumn{4}{c}{
CAMO}&\multicolumn{4}{c}{
COD10K}\\\cline{3-14}
& & \small{$M\downarrow$} & \small{$F_{\beta}\uparrow$} & \small{$E_{\phi}\uparrow$} &  \small{$S_{\alpha}\uparrow$}& \small{$M\downarrow$} & \small{$F_{\beta}\uparrow$} & \small{$E_{\phi}\uparrow$} &  \small{$S_{\alpha}\uparrow$} & \small{$M\downarrow$} & \small{$F_{\beta}\uparrow$} & \small{$E_{\phi}\uparrow$} &  \small{$S_{\alpha}\uparrow$} \\
\hline
\multicolumn{14}{c}{Scribble Supervision Setting} \\
\cline{1-14}
WSSA\cite{zhang2020weakly} & \small{CVPR20} & 0.067 & 0.692 & 0.860 & 0.782 & 0.118 & 0.615 & 0.786& 0.696 & 0.071 & 0.536 & 0.770 & 0.684 \\
SCWS\cite{yu2021structure} & \small{AAAI21} & 0.053 & 0.758 & 0.881 & 0.792 & 0.102 & 0.658 & 0.795 & 0.713 & 0.055 & 0.602 & 0.805 & 0.710\\
TEL\cite{zhang2020weakly} & \small{CVPR22} & 0.073 & 0.708 & 0.827 & 0.785 & 0.104 & 0.681 & 0.797 & 0.717 & 0.057 & 0.633 & 0.826 & 0.724 \\
SCOD\cite{he2023weakly} & \small{AAAI23} & \textbf{0.046} & \textbf{0.791} & \textbf{0.897} & \textbf{0.818} & \textbf{0.092} & \textbf{0.709} & \textbf{0.815} & \textbf{0.735} & 0.049 & 0.637 & \textbf{0.832} & 0.733 \\
SAM\cite{kirillov2023segment} & \small{ICCV23} & 0.207 & 0.595 & 0.647 & 0.635 & 0.160 & 0.597 & 0.639 & 0.643 & 0.093 & 0.673 & 0.737 & 0.730 \\
SAM-S\cite{kirillov2023segment} & \small{ICCV23} & 0.076 & 0.729 & 0.820 & 0.650 & 0.105 & 0.682 & 0.774 &0.731 & \textbf{0.046} & \textbf{0.695} & 0.828 & \textbf{0.772} \\
\hline
\multicolumn{14}{c}{Point Supervision Setting} \\
\cline{1-14}
WSSA\cite{zhang2020weakly} & \small{CVPR20} & 0.105 & 0.660 & 0.712 & 0.711 & 0.148 & 0.607 & 0.652 & 0.649 & 0.087 & 0.509 & 0.733 & 0.642 \\
SCWS\cite{yu2021structure} & \small{AAAI21} & 0.097 & 0.684 & 0.739 & 0.714 & 0.142 & 0.624 & 0.672 & \textbf{0.687} & 0.082 & 0.593 & 0.777 & 0.738\\
TEL\cite{zhang2020weakly} & \small{CVPR22} & 0.094 & \textbf{0.712} & \textbf{0.751} & \textbf{0.746} & 0.133 & \textbf{0.662} & 0.674 & 0.645 & 0.063 & 0.623 & \textbf{0.803} & 0.727 \\
SCOD\cite{he2023weakly} & \small{AAAI23} & \textbf{0.092} & 0.688 & 0.746 & 0.725 & 0.137 & 0.629 & 0.688 & 0.663 & \textbf{0.060} & 0.607 & 0.802 & 0.711 \\
SAM\cite{kirillov2023segment} & \small{ICCV23} & 0.207 & 0.595 & 0.647 & 0.635 & 0.160 & 0.597 & 0.639 & 0.643 & 0.093 & 0.673 & 0.737 & 0.730 \\
SAM-P\cite{kirillov2023segment} & \small{ICCV23} & 0.101 & 0.696 & 0.745 & 0.697 & \textbf{0.123} & 0.649 & \textbf{0.693} & 0.677 & 0.069 & \textbf{0.694} & 0.796 & \textbf{0.765}\\\hline
\multicolumn{14}{c}{Task-Generic Prompt Setting} \\
\cline{1-14}

{\small CLIP Surgery+SAM}\cite{li2023clip} & \small{Arxiv2023} &  0.180 & 0.557 & 0.710 & 0.637& 0.206& 0.466& 0.666 & 0.573 & 0.187 & 0.448 &0.672 & 0.601\\
\rowcolor{gray!20}GenSAM & \small{Ours} & \textbf{0.090} & \textbf{0.680} & \textbf{0.807} & \textbf{0.764} & \textbf{0.113} & \textbf{0.659} & \textbf{0.775} & \textbf{0.719} &\textbf{0.067} & \textbf{0.681} & \textbf{0.838} & \textbf{0.775} \\
\hline
\end{tabular}}
\end{table*}
We also obtain the corresponding background consensus $SI_{back}$ in a similar way.
%
Then the resulting similarity heatmap $SI$ is:
\begin{equation}
    SI = SI_{fore} - SI_{back},
\end{equation}
where $SI \in \mathbb{R}^{N_i \times 1}$. 
$SI$ is then upsampled using bilinear interpolation. After upsampling $SI$ to match the original size of image $X$, the resulting output can be regarded as the consensus heatmap $H$ corresponding to $X$.
 \jy{We further sample highly-activated pixels on $H$ as positive point prompts and the same number of the most unactivated pixels as negative point prompts
 to guide the segmentation process in SAM.}
 
\subsection{Progressive Mask Generation} 
 \jy{However, a single inference
 may not provide satisfactory segmentation result.
 For image with complicated background, 
 some backgound objects can also be highly activated in the heatmap, causing some noises for inference the point prompts.
 In order to get more roubust prompt,}
 we use the heatmap as a visual prompt to reweight the original image and guide the model during test time adaptation. 
The weighted image $X'$ is as follows:
\begin{align}
    X' = X * H * w_{pic} + X * (1 - w_{pic}),
    \label{eq:update}
\end{align}
where $w_{pic}=0.3$ is a hyper-parameter.
The weighted image $X'$ is then used as input image for next iteration.
In this way, we  develop a circularly test-time adaptation framework which involves multiple iterations of inference in a coarse-to-fine manner. 
%
%
%

Moreover, in subsequent iterations, we use the previous iteration's mask to guide the segmentation by drawing bounding boxes as a post-process. 
We select the box with the highest Intersection over Union (IoU) value with the mask as our choice. 
It optimizes the current iteration and improves the consistency of segmentation results.
The mask obtained at the $i$-th iteration is defined as $M_i$, where $i \in \{1,.., \mathbf{Iter}\}$, $\mathbf{Iter}$ is set as 6.
To eliminate the impact of ambiguity caused by inconsistent prompts in each iteration, the mask obtained in each iteration is averaged. 
Finally, the selected iteration $\mathrm{i}^*$ is determined by selecting the iteration's result that closely resembles the average mask across all iterations as follows:
\begin{small}
\begin{equation}
\mathrm{i}^* = \arg\min_{i}\left(\left| M_i - \frac{\sum_{i}{ (M_{1}, \ldots, M_{\mathbf{Iter}}})}{\mathbf{Iter}} \right|\right),
\end{equation}
\end{small}
then $M_{\mathrm{i}^*}$ is the corresponding final mask for $X$.
\section{Experiments}

\begin{table*}[tb!]
    \caption{Ablation study of variants with our GenSAM on camouflaged object detection.}
    \centering 
  \resizebox{1.0\textwidth}{!}{
    \begin{tabular}{ccccc|cccc|cccc|cccc}
        \hline
         \multicolumn{5}{c}{\multirow{2}{*}{\centering{Method's variant}}}  & \multicolumn{12}{|c}{settings on camouflaged object detection}\\ \cline{6-17}  
     & & & & &\multicolumn{4}{c|}{CHAMELEON} &\multicolumn{4}{c|}{CAMO}&\multicolumn{4}{c}{COD10K}\\ \hline
     BLIP2 keyword & chain foreground&PMG&kkv self-attention&chain background& \small{$M\downarrow$} & \small{$F_{\beta}\uparrow$} & \small{$E_{\phi}\uparrow$} &  \small{$S_{\alpha}\uparrow$}& \small{$M\downarrow$} & \small{$F_{\beta}\uparrow$} & \small{$E_{\phi}\uparrow$} &  \small{$S_{\alpha}\uparrow$} & \small{$M\downarrow$} & \small{$F_{\beta}\uparrow$} & \small{$E_{\phi}\uparrow$} &  \small{$S_{\alpha}\uparrow$} \\
        \hline 
     & & & & & 0.180 & 0.557 & 0.710 & 0.637& 0.206& 0.466& 0.666 & 0.573 & 0.187 & 0.448 &0.672 & 0.601 \\
       \checkmark& & & & &
       0.106 & 0.689 & 0.803 & 0.749 &
       0.200& 0.503& 0.676 & 0.602 & 
        0.146&0.556&0.735&0.673 \\
       \checkmark&\checkmark& & & &
       0.094 & 0.687 & 0.800 & 0.754 & 0.198 & 0.521 &0.687 &
       0.613& 0.143 &0.569 & 0.740& 0.681 \\
       \checkmark&\checkmark&\checkmark & & &
       0.098 &  0.659 & 0.779 & 0.741& 0.161 & 0.554  & 0.719 & 0.642 & 0.086 &0.616 & 0.797&0.731\\
       \checkmark&\checkmark&\checkmark & \checkmark&& 
       \textbf{0.078} & \textbf{0.711} & \textbf{0.817} & \textbf{0.776} & 0.147&0.583& 0.746 & 0.666 & 0.069 & 0.660 &
       0.820 & 0.760 \\
    
    \rowcolor{gray!20}\checkmark&\checkmark&\checkmark&\checkmark&\checkmark& 
    0.090& 0.680&0.807 & 0.764 &\textbf{0.113}& \textbf{0.659}&	\textbf{0.775}&\textbf{0.719}&\textbf{0.067}& \textbf{0.681}& \textbf{0.838}& \textbf{0.775}\\\hline
    \end{tabular}%
    }
\label{table:component}
\end{table*}

\begin{table*}[tb!]

\caption{{Ablation study on COD10K. }}
\subcaptionbox{\scriptsize{Number of chains}.
     \label{fig:chain}}{
     \begin{minipage}[t]{0.17\linewidth}
        \centering
        \tiny
        \setlength{\tabcolsep}{1pt}
        \begin{tabular}{c|cccc} 
        \hline
        Chains      & {$M\downarrow$} & {$F_{\beta}\uparrow$} & {$E_{\phi}\uparrow$} &  {$S_{\alpha}\uparrow$}\\ \hline
        1  & 0.069          & 0.671          & 0.827          & 0.767          \\
        2  & \textbf{0.066}          & 0.679          & 0.832          & 0.772          \\
        \rowcolor{gray!20}3  & {0.067} & 0.681 & \textbf{0.838} & \textbf{0.775}\\
        4  &\textbf{0.066} &0.680  &0.834 &  0.773 \\
        5  & \textbf{0.066}       & \textbf{0.683} & 0.833    &  \textbf{0.775}\\ \hline
    \end{tabular}
 \end{minipage}}
    \subcaptionbox{\scriptsize{Heatmap upsample factor}.
       \label{fig:scale}}{
     \begin{minipage}[t]{0.17\linewidth}
        \centering
        \tiny
         \setlength{\tabcolsep}{1pt}
        \begin{tabular}{c|cccc} 
        \hline
        Factor      &  {$M\downarrow$} & {$F_{\beta}\uparrow$} & {$E_{\phi}\uparrow$} &  {$S_{\alpha}\uparrow$} \\ \hline
        8  & 0.110 & 0.496 & 0.750 & 0.689  \\
        4   & 0.082 & 0.596 & 0.806  & 0.741  \\
        \rowcolor{gray!20}2   & \textbf{0.067} & \textbf{0.681}    & \textbf{0.838}   & \textbf{0.775} \\
        1      & 0.081 & 0.658 & 0.808  & 0.754  \\
        0.5      & 0.107 & 0.595 & 0.753  & 0.708  \\ \hline
    \end{tabular}
 \end{minipage}}
 \subcaptionbox{\scriptsize{Heatmap threshold}.
       \label{fig:threshold}}{
     \begin{minipage}[t]{0.17\linewidth}
        \centering
         \renewcommand{\arraystretch}{1.2}
        \tiny
         \setlength{\tabcolsep}{1pt}
        \begin{tabular}{c|cccc} 
        \hline
        Thr.      &{$M\downarrow$} & {$F_{\beta}\uparrow$} & {$E_{\phi}\uparrow$} &  {$S_{\alpha}\uparrow$} \\ \hline
        0.80        & 0.107 & 0.549 & 0.767  & 0.717  \\
        0.85       & 0.080 & 0.623 & 0.814  & 0.754  \\
        \rowcolor{gray!20}0.90        & \textbf{0.067} & \textbf{0.681}    & \textbf{0.838}   & \textbf{0.775}                               \\
        0.95       & 0.068 & 0.679 & 0.818  & 0.763  \\\hline
        \end{tabular}
 \end{minipage}}
 \subcaptionbox{\scriptsize{Post processing.}  \label{fig:Postprocessing}}{
 \begin{minipage}[t]{0.22\linewidth}
    \centering
    \renewcommand{\arraystretch}{1.2}
    \tiny
    \setlength{\tabcolsep}{1pt}
    \begin{tabular}{c|cccc} 
    \hline
    Post processing      & {$M\downarrow$} & {$F_{\beta}\uparrow$} & {$E_{\phi}\uparrow$} &  {$S_{\alpha}\uparrow$} \\ \hline
    None       & 0.073 & \textbf{0.683}  & 0.822  & 0.763  \\
    MaxBox     & 0.107 & 0.639 & 0.799  & 0.746  \\
    Mask       & 0.114 & 0.666 & 0.800  & 0.753  \\
    \rowcolor{gray!20}MaxIOUBox  & \textbf{0.067} & 0.681   & \textbf{0.838}   & \textbf{0.775}                               \\\hline
    \end{tabular}
 \end{minipage}}
 \subcaptionbox{\scriptsize{Iteration of adaptation. }\label{fig:iteration}}{
 \begin{minipage}[t]{0.22\linewidth}
 \includegraphics[height=1.6cm]{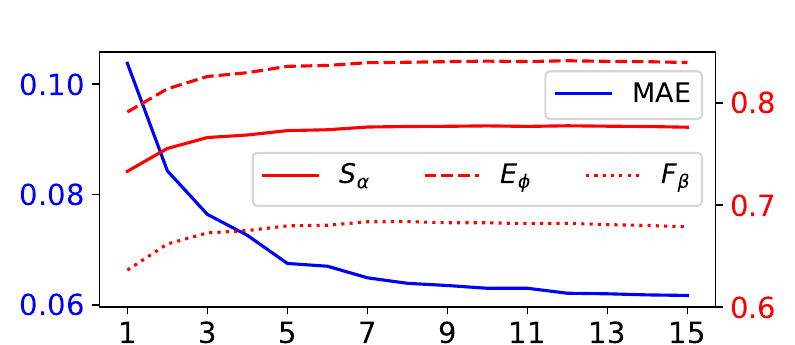}
 \end{minipage}}
 \label{fig:abla}
\end{table*}

To evaluate GenSAM in different scenarios, we  choose challenging camouflaged object detection (COD) datasets to evaluate our GenSAM under various settings.
%

\subsection{Setup}
\noindent\textbf{Datasets.} Camouflaged object detection tasks aim to identify organisms attempting to camouflage themselves from complex backgrounds. 
In this study, we select three representative datasets containing samples of camouflage objects: CHAMELEON \cite{skurowski2018animal}, CAMO \cite{le2019anabranch} and COD10K \cite{fan2021concealed}. 
CHAMELEON dataset comprises 76 images sourced from the Internet for testing purposes. 
CAMO dataset consists of 1,250 images, with 1,000 images allocated for training and 250 images for testing. 
COD10K dataset 
%
encompasses a total of 3,040 training samples and 2,026 testing samples.

\noindent\textbf{Baseline.}
We  compare current SOTA weakly supervised segmentation methods, namely SAM \cite{kirillov2023segment}, WSSA \cite{zhang2020weakly}, SCWS \cite{yu2021structure}, TEL \cite{zhang2020weakly}, and SCOD \cite{he2023weakly}.
Three distinct levels of supervision are introduced, including scribble supervision and point supervision, alongside our proposed task-generic prompt settings.
Following the previous weakly supervised segmentation setting \cite{he2023weakly1}, in terms of scribble supervision, it involves providing foreground and background supervision by drawing the primary structure of objects and background areas. 
Point supervision refers to the provision of separate points as supervision for the foreground and background.
In our newly proposed generic task prompt, we provide a unified prompt description ``the camouflaged animal'' for all images. 
The model is required to independently convert this unified description into specific supervision to guide the segmentation process based on the characteristics of each image.
Regarding SAM, we follow the suggested setup by \cite{he2023weakly1} that involves comparing two variants of SAM: SAM-S and SAM-P. They finetune the mask decoder of SAM through scribble and point supervision training data respectively. 
Both variants employ partial cross-entropy loss for training.
Note that all the comparative methods we  employ in our study are trained on camouflage segmentation datasets and tested on a separate test set. 
However, GenSAM does not require training data at all, while directly utilize the test set for test-time adaptation.
Based on the approaches used in previous studies \cite{fan2021concealed, fan2020camouflaged}, we employ four commonly used metrics for evaluation, include Mean Absolute Error ($M$), adaptive F-measure ($F_{\beta}$) \cite{margolin2014evaluate}, mean E-measure ($E_{\phi}$) \cite{fan2021cognitive}, and structure measure ($S_{\alpha}$) \cite{fan2017structure}. 
It is worth noting that a smaller value of $M$ or larger values of $F_{\beta}$, $E_{\phi}$, and $S_{\alpha}$ indicate better segmentation performance.

\noindent\textbf{Implementation Details.}
For image-to-caption model, we use the BLIP-2 ViT-g OPT$_{6.7B}$ verison of BLIP2. For CLIP, we apply the CS-ViT-B/16 pretrained model. As for obtaining the point prompt from the heatmap, we use threshold=0.9 to filter out all the positive points and get the same amount of points with the lowest scores as the negative point prompt.
In each iteration, we use the output mask generated from the last iteration as an auxiliary prompt in addition to the point prompts to guide SAM in the current iteration, to ensure consistency in each iteration.
We totally apply 6 iterations to get our best results.
We use PyTorch framework and conduct experiments on a single NVIDIA A100 GPU.
\subsection{Experiment Results and Analysis}
\begin{figure}[tb]
  \centering
  \includegraphics[width=8.4cm]
{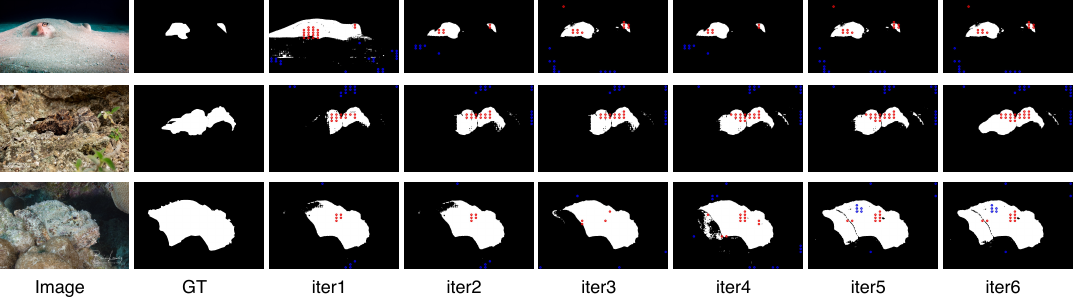}   
   \caption{Iterative qualitative results of GenSAM.
   The visualized results indicate that as test-time adaptation progresses, the segmentation results consistently improve.}\label{fig:qualitative}
  \end{figure}

\noindent\textbf{Experiment Results.}
As shown in Tab. \ref{tab:results}, we compare GenSAM with approaches that use different supervision methods, including scribble supervision, point supervision, and our newly proposed generic task prompt supervision.
Overall, due to varying levels of supervision signals, scribble supervision outperformed point supervision. 
However, our GenSAM, despite having only one generic task prompt as the universal supervision for the entire dataset, achieved superior performance compared to point supervision in terms of the $M$, $S_{\alpha}$, and $E_{\phi}$ metrics on CHAMELEON. 
GenSAM approaches the performance level of scribble supervision. This trend is even more pronounced on the more challenging CAMO and COD10K datasets. 
Remarkably, on the most challenging COD10K dataset, our method even achieves better results in terms of the $S_{\alpha}$ and $E_{\phi}$ metrics under the less supervised generic task prompt supervision, surpassing both scribble supervision and point supervision approaches. 
This further demonstrates the superiority of GenSAM. Additionally, our method consistently outperforms SAM, SAM-P, SAM-S and CLIP Surgery+SAM, indicating that the improvements of our method stem from its own merits rather than relying solely on the superior segmentation capabilities of SAM. The qualitative results are shown in Fig. \ref{fig:qualitative}. 

\noindent\textbf{Component Analysis}
We further analyze the impact of components in Tab. \ref{table:component}. When all modules are removed, the model became CLIP Surgery+SAM. We use the general task description ``the camouflaged animal'' for each sample. The performance is weak across various datasets in this case. This result emphasizes the effectiveness of our complete GenSAM approach.
In the third row, we add heatmaps as visual prompts during the iterative process. It is noticed that there is a notable performance improvement. It shows the significance of setting heatmaps as visual prompts, although there is still performance gap compared to GenSAM.
In the second and fifth rows, we add consensus heatmaps for foreground and background using the chain-of-thought prompting. The comparison with other experiments emphasizes the importance of chain-of-thought for achieving consensus. 
The last two rows emphasize the importance of using a chain-of-thought to remove background interference.
The unusual results on the CHAMELEON dataset are due to its small size, resulting in greater randomness, while the results from the other two larger and more complex datasets match our expectations.
In the forth row, we replace our k-k-v self-attention with the v-v-v self-attention from CLIP Surgery. A notable decrease in performance is observed compared to using k-k-v, indicating the impact of k-k-v self-attention.
%
As shown in Fig.~\ref{fig:iteration},
%
the performance of the model's test-time adaptation shows a significant boost within the first 1-6 iterations. Then it gradually stabilizes thereafter. 
Although the best performance is achieved at the 8th iteration, the improvement compared to the 6th iteration is not substantial, and it incurs additional time loss. 
Therefore, we set the number of iterations to 6.
%
\begin{table}[tb!]
\caption{Results on Polyp Image Segmentation and Shadow Detection with generic task prompt.}
 \centering
    \resizebox{1.0\columnwidth}{!}{
\begin{tabular}{l|l|cccc}
\hline
Datasets & Methods     & M$\downarrow$ & \small$F_{\beta}\uparrow$ & $E_{\phi}\uparrow$ & $S_{\alpha}\uparrow$ \\ \hline
\multirow{2}{*}{\begin{tabular}[l]{@{}l@{}} ETIS~\cite{silva2014toward}\\      (Polyp Image Segmentation)\end{tabular}} & CLIP Surgery+SAM &  0.537 & 0.047 & 0.296 & 0.272 \\
 & GenSAM & \textbf{0.205}   & \textbf{0.090} & \textbf{0.554}  & \textbf{0.430}  
 \\\hline

\multirow{2}{*}{\begin{tabular}[l]{@{}l@{}}SBU\cite{vicente2016large} \\      (Shadow Detection)\end{tabular}}    & CLIP Surgery+SAM & 0.331 & 0.336 & 0.517 & 0.442 \\
 & GenSAM & \textbf{0.215}   & \textbf{0.421}   & \textbf{0.621}  & \textbf{0.529} \\ \hline      
\end{tabular}
}
    \centering
    \label{tab:generalization}
\end{table}

\noindent\textbf{Generalization of GenSAM.}
In Tab. \ref{tab:generalization}, we assess GenSAM's performance on two segmentation tasks: Polyp Image Segmentation and Shadow Detection. We employ the generic prompts ``Polyp'' and ``Shadow" for them respectively. Experiments demonstrate the significant improvement achieved by our method compared to the baseline.

\noindent\textbf{Number of chains.}
We also evaluate the impact of the number of chains in the chain-of-thought prompting in Tab. \ref{fig:chain}.
When the number of chains is equal to or less than 3, performance gradually improves with an increasing number of chains. 
However, when the number of chains exceeded 3, although the inference time increases, there is no significant improvement in performance or even a decline. 

\noindent\textbf{From heatmap to point prompt.} In Tab.~\ref{fig:scale}-\ref{fig:threshold} we performe a parameter scan of heatmap upsample factor, the threshold to get the point prompts from heatmap. As the original heatmap $SI$ is of relatively low resolution ($14\times14)$, we upsample  $SI$ to obtain $H$ and obtain the point prompt using a threshold. We finally set the  heatmap upsample factor as 2 and the threshold as 0.9

\noindent\textbf{Generating box prompt.} 
In Tab.~\ref{fig:Postprocessing}, we try different methods to transform the mask output from the previous iteration into an auxiliary mask or box prompt in addition to the point prompt to guide the current iteration, which ensures the consistency of mask outputs in different iterations.
We try different transformation methods, including directly using the last mask output as the mask prompt (Mask), using the maximum surrounding box (MaxBox) and the box that has the largest Intersection over Union of the mask (MaxIOUBox).  Results show that MaxIOUBox outperforms other strategies 
\section{Conclusion}
In this paper, we present GenSAM, which automatically generates image-specific consensus prompts for WSCOD with only a generic task description via our test-time progressive mask generation framework. Experiments on various COD datasets show GenSAM's superiority.

\noindent\textbf{Acknowledgements.} This work was supported by Veritone, the Alan Turing Institute Turing Fellowship, and the China Scholarship Council.
\bibliography{aaai24}

\end{document}